\documentclass[5p,times]{elsarticle}

\usepackage{amssymb}
\usepackage{algorithm}
\usepackage{algorithmic}
\usepackage{subfigure}

\usepackage{amsmath}
\usepackage{amssymb} 
\usepackage{mathtools}
\usepackage{xcolor}
\usepackage{natbib}
\setcitestyle{numbers,square}

\begin{document}

\begin{frontmatter}



\author{Yajie Zhang\fnref{label2,label3}}
\ead{zyj0928@tju.edu.cn}
\fntext[label2]{College of Intelligence and Computing, Tianjin University, No.135 Yaguan Road, Haihe Education Park, Tianjin 300350, China}
\fntext[label3]{Technical $R\&D$ Innovation Center, National Astronomical Data Center, No.135 Yaguan  Road, Haihe Education Park, Tianjin 300350, China}

\author{Ce Yu\fnref{label2,label3}\corref{cor1}}
\cortext[cor1]{Corresponding author.}
\ead{yuce@tju.edu.cn}

\author{Chao Sun\fnref{label2,label3}}

\author{Jizeng Wei\fnref{label2,label3}}

\author{Junhan Ju\fnref{label2,label3}}

\author{Shanjiang Tang\fnref{label2,label3}}

\title{Solving Online Resource-Constrained Scheduling for Follow-Up Observation in Astronomy: a Reinforcement Learning Approach}






\begin{abstract}

\end{abstract}

\begin{abstract}
In the astronomical observation field, determining the allocation of observation resources of the telescope array and planning follow-up observations for targets of opportunity (ToOs) are indispensable components of astronomical scientific discovery. This problem is computationally challenging, given the online observation setting and the abundance of time-varying factors that can affect whether an observation can be conducted. This paper presents \texttt{ROARS}, a reinforcement learning approach for online astronomical resource-constrained scheduling. To capture the structure of the astronomical observation scheduling, we depict every schedule using a directed acyclic graph (DAG), illustrating the dependency of timing between different observation tasks within the schedule. Deep reinforcement learning is used to learn a policy that can improve the feasible solution by iteratively local rewriting until convergence. It can solve the challenge of obtaining a complete solution directly from scratch in astronomical observation scenarios, due to the high computational complexity resulting from numerous spatial and temporal constraints. A simulation environment is developed based on real-world scenarios for experiments, to evaluate the effectiveness of our proposed scheduling approach. The experimental results show that \texttt{ROARS} surpasses 5 popular heuristics, adapts to various observation scenarios and learns effective strategies with hindsight.

\end{abstract}




\end{frontmatter}

\section{Introduction}
Astronomical sky surveys, the primary avenue for exploring the universe, have generated numerous scientific breakthroughs \cite{ma2020night,santana2022orbital}. Observation scheduling constrained by time-varying observation conditions and shared limited resources is a crucial problem for survey observation, using a telescope array with multiple telescopes. The visibility of celestial objects changes in real time and the lifetime of expensive astronomical observation equipment is limited \cite{Zhang_2023}. So efficient observation scheduling and resource management are conducive to maximizing scientific output. Making follow-up observations for targets of opportunity (ToOs) in sky surveys refers to gathering additional data to further understand specific transient phenomena of great scientific interest identified during the initial survey \cite{liu2018research}. It requires the use of various instruments and specific filters as astronomers are committed to unveil more details, enhancing the comprehensive exploration of the cosmos.

\begin{figure*}[t]
\centering
\includegraphics[width=0.5\textwidth]{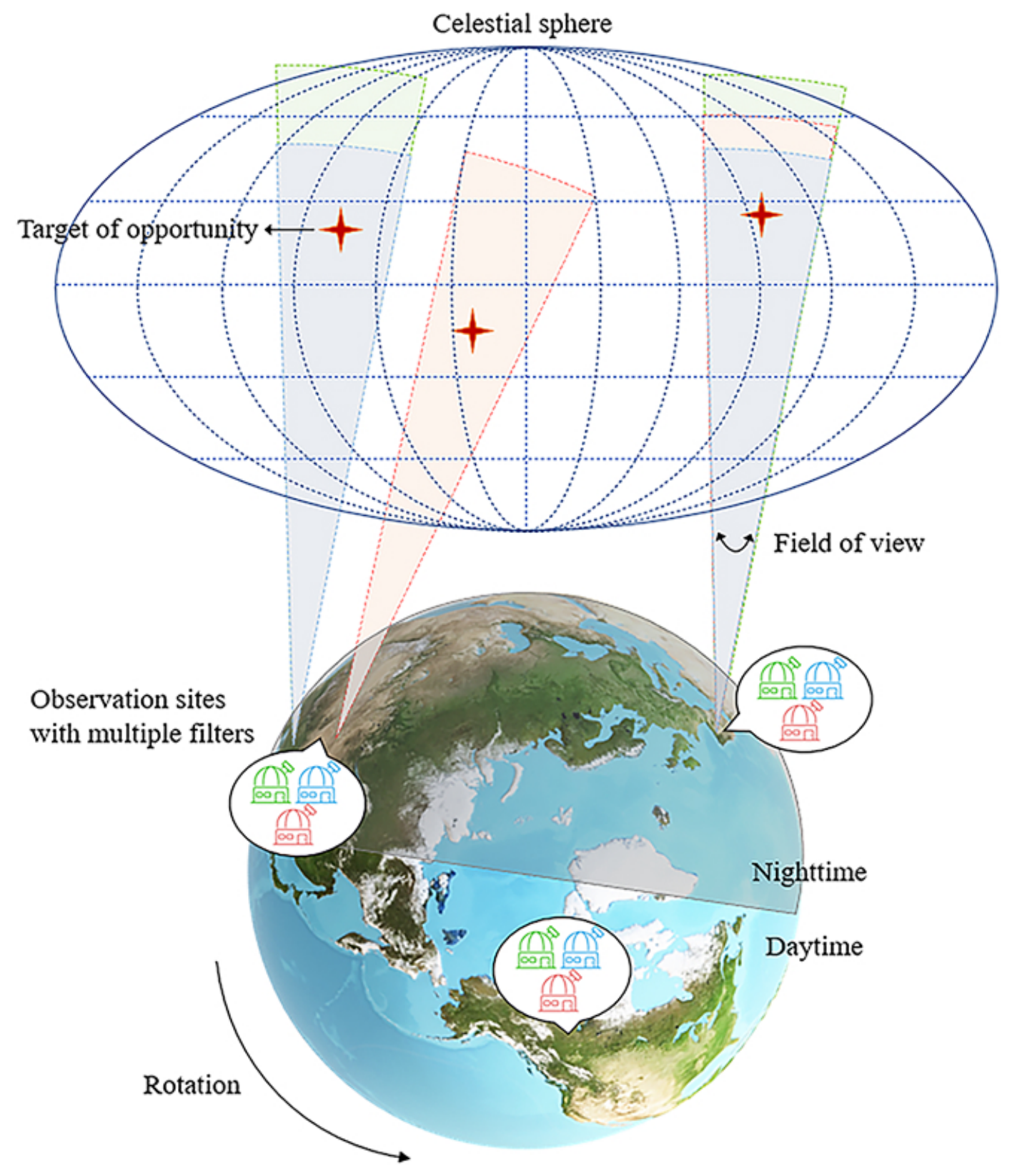} 
\caption{Conceptual illustration of distributed telescope array follow-up observation for targets of opportunity. } 
\label{observation schema}
\end{figure*}

The complexity of the astronomical observation task scheduling problem has been proved to be NP-hard \cite{bonvallet2010methodological}. The difficulty of solving resource-constrained task scheduling for online follow-up observations optimally is rooted in its intricate nature. In the process of time-domain sky survey, it is a kind of closed-loop scheduling to deal with ToOs that appear suddenly. The multi-band follow-up scheduling algorithm needs to deal with the uncertainty of resource availability (including observable time limited by observation conditions, telescope filters, etc.), target properties (arrival time, duration, and requirements for observation band, exposure time and observation mode), etc. Note that these target priority constraints are determined by astronomers for scientific discovery after the target appears, and there are no prior probability distributions. Specifically, incoming follow-up observation targets have various precedence relations and observation band requirements, often requiring simultaneous observations by multiple telescopes with different filters across bands, as illustrated in Fig. \ref{observation schema}. As the Earth rotates, the visibility of each target changes over time. The observation time window of the target is affected by the positions of the observation sites and the target and changes with time, which is an important computational challenge in our scheduling problem. Through observation sites that distributed around the world, different ToOs can be coordinated observation. Careful consideration of execution order and time-varying constraints is essential for coordinated and timely observations. In distributed telescope array environment, the variance in available observation time across sites, coupled with potential competition for target observations due to their distribution, presents greater challenges to the execution of observation plans \cite{jia2023observation}. The majority of these issues are addressed using integer linear programming (ILP) and meticulously designed heuristics. For example, the Zwicky Transient Facility allocates survey fields to time blocks, minimizing the need for filter changes and ensuring precise control over the number of exposures per field \cite{bellm2019zwicky}. Similar optimization objectives are considered for the Large Synoptic Survey Facility using genetic algorithms \cite{naghib2019framework}. Moreover, heuristic methods are widely employed for solving this problem within the astronomical field, e.g., lowest airmass  \cite{rana2019optimal}, smallest telescope slew angle \cite{rana2017enhanced}, for finding optical counterparts to transients. However, for practical large-scale survey observation, determining when and where to apply these heuristics, and establishing their prioritization, is both inflexible and time-consuming. In addition, meta-heuristic algorithm has been used for agile earth observation satellite scheduling problem \cite{wu2023frequent}, but there are significant differences in the observation modes of satellites and telescopes.

Revisiting the above challenges, given the problem scale and computational complexity, we investigate the use of reinforcement learning (RL) approaches for this real-world resource-constrained scheduling problem. RL benefits from abundant training data generated by repetitive scheduling decisions in astronomical observation systems, allowing for effective learning. Meanwhile, RL has the capability to model intricate systems and decision-making strategies through deep neural networks, akin to the architectures employed for gaming agents \cite{silver2016mastering}. This is achieved by integrating various input signals to enable online application in stochastic environments \cite{mao2016resource}. However, the application of RL-based scheduling to astronomical observation domains is not straightforward given the lack of well-suited model and benchmark. The processing of high-priority follow-up tasks and the efficient generation of observation plans during the survey are essential to the guarantee of astronomical discoveries. Therefore, as the key to developing effective observation resource allocation plans, the effective acquisition astronomical observation knowledge and observation modeling are pressing issues that demand urgent attention.

To our knowledge, the combination of online resource management, astronomical follow-up ToO observation and telescope array is not considered in the literature so far. So in this paper we contribute by, first, defining and modeling the real-world resource-constrained scheduling problem in telescope array setting, considering both multiple telescopes in one observation site (we call intra-site) and geographically distributed in multiple sites. Second, we propose a new approach to tackle the problem using deep reinforcement learning (DRL), named \texttt{ROARS}, for a Reinforcement learning approach for Online Astronomical Resource-Constrained Scheduling. A directed acyclic graph (DAG) is used to extracts the problem-specific knowledge and model the temporal dependency of the problem. High quality solutions with different sizes and structures are learned by iteratively refining \cite{chen2019learning} existing solutions towards optimality using DRL. Because the approach of obtaining a complete solution directly from scratch, as employed in previous similar works \cite{mao2016resource,vinyals2015pointer,kool2018attention}, becomes challenging to achieve feasibility when scaling up the observation scheduling. Here we focus on online setting where the follow-up observation tasks arrive dynamically with unpredictable constrains and cannot be preempted once scheduled. The proposed \texttt{ROARS} is evaluated through extensive simulations performed on real-world data. Our preliminary results show that \texttt{ROARS} is able to generate solutions of consistent quality in various astronomical observation scenarios, thereby facilitating robust and rapid schedule adaptation amidst uncertainty. The quality of schedules generated by \texttt{ROARS} can exceed those of the heuristic baselines and offline scheduling scenarios, where the entire observation task sequence is known prior to scheduling. The approach can also be extended to distributed telescope array observation environments with robustness performance.

\section{Related Work}

Resource scheduling and optimization problems, pervasive and fundamental issues in complex system, have been extensively studied from both theoretical and empirical perspectives \cite{hartmann2022updated,liu2023late}. There are numerous studies demonstrating effective use of DRL in several real-world resource scheduling scenarios, such as smart manufacturing \cite{wang2021dynamic}, city-wide firefighting \cite{iqbal2022alma}, steel production \cite{zhou2023reinforcement,feng2024gas}, resource provisioning in Internet of Things ecosystem \cite{chowdhury2019drls, tran2022reinforcement}, logistics and retail \cite{zhou2023reinforcement}. Existing studies demonstrate that RL exhibits effectiveness in terms of the solution quality, and can achieve substantial time savings compared to the classical heuristic approaches \cite{mazyavkina2021reinforcement}. Therefore, DRL is extensively investigated as an effective approach for controlling complex systems.

Despite the clear need, there is an absence of research undertaken in the area of intelligent astronomical observation resource allocation for telescope array. Nowadays, resource allocation and management methods in most astronomical survey observation projects can be divided into two types, based on ILP algorithms \cite{lampoudi2015integer, solar2016scheduling} and human-generated heuristics \cite{liu2018research}. Nevertheless, with the increase of the number of telescopes and observation targets, especially in the environment of telescope array observation, fine-scale observation strategy optimization requires extensive manual intervention, which exceeds the ability of conventional planning algorithms and classical solvers. For the observation environment using a distributed telescope array, a flexible multilevel global scheduling model is proposed for a generic telescope array scheduling problem by Zhang et al. \cite{Zhang_2023}. While their algorithm produces long-term scheduling solutions in survey observation mode, it does not undertake precise resource coordination for follow-up observations of ToOs. Jia et al. implements a telescope array observation simulator and applies DRL into a space debris observation scenario \cite{jia2023observation}. But since publication, there is currently no established general approach for resource management in online follow-up astronomical observation using an array of multiple telescopes.

The follow-up observation scheduling problem in astronomy can be seen as a special resource-constrained project scheduling problems (RCPSP \cite{demeulemeester2002project}). In order to achieve the robust scheduling \cite{lambrechts2008tabu,lambrechts2011time}, researchers propose multiple heuristic and meta-heuristic procedures to allocate time buffers in a given schedule while ensuring adherence to a predefined project due date \cite{van2008proactive}. By contrast, inserting time buffers in a proactive way to deal with scheduling uncertainties is not in line with the principle of telescopic observation, because the telescope is expensive and has limited life, observation time is a very valuable resource. Li et al. develop efficient approximate dynamic programming (ADP) algorithms for RCPSP with uncertain task duration, using constraint programming and a hybrid ADP framework to enhance performance and efficiency \cite{li2015solving}. Brvcic et al. address the issue of inflexibility in proactive–reactive scheduling by introducing threshold-based cost functions for deviation penalties in projects with stochastic task duration \cite{brvcic2019planning}. While Xie et al. focus on RCPSP with uncertain resource availability \cite{xie2021approximate}, they use a new Markov decision process model and a rollout-based ADP algorithm, significantly improving performance over heuristic methods. Compared with these traditional problems, the problem of RCPSP in time-domain survey to be solved in this paper focuses more on fast processing of special targets to ensure their observation quality, rather than maintaining the stability of the original tasks.

With the development of deep learning technology in recent years, one research \cite{mao2016resource} presents an example solution that transforms the problem of packing tasks with diverse resource demands into a learning problem. The resource allocation strategies are directly learned from experience. However, it only considers a single-cluster situation, and factors such as the dependency between jobs have not been investigated. Another research \cite{teichteil2023fast} relies on the graph neural network to address RCPSP of varying sizes, including in presence of uncertain task duration. Cai et al. further solves the RCPSP with resource disruptions, and uses proximal policy optimization (PPO) to train the model in an end-to-end way for performance optimization \cite{cai2024deep}. Their work is relevant for us, but for the follow-up observation scenario in astronomical domain, scheduling strategies should consider real locations of telescopes, distributions of observation targets, and filter requirements. The constrained observation conditions are closely related to these time-varying factors. Therefore, the current industrial scheduling methods are difficult to be directly applied in the field of astronomical observation.

In other recent work, the local rewriting is proposed for combinatorial optimization \cite{chen2019learning}, the performance is assessed across three distinct domains: online job scheduling, expression simplification, and vehicle routing. It has shown better performance than heuristics using multiple metrics in solving complex problems where generating an entire solution directly is challenging. Given its effectiveness in capturing hierarchical and sequential structures, and order constraints \cite{yu2020reinforcement}, the Child-Sum Tree-LSTM architecture is well-suited for the dynamic and complex nature of resource scheduling in astronomical observations. So it is clear that despite a lack of exploration into a general intelligent resource management approach in the astronomical observation domain, the existence of mature and extensive research supports the exploration as a feasible approach for tackling the application challenges.

\section{Follow-up observation scheduling in astronomy}
\label{notions}

In this section, the problem setting and parameters are presented first, followed by its MDP formulation, which lays the foundation for our ROARS algorithm to be developed in the succeeding sections.

\subsection{Problem statement}

\begin{table*}[htbp]
	\centering
\caption{\textbf{Summary of key parameters and variables.}}
	\begin{tabular}{|c|l|}
		\hline 
		\textbf{Notation} & \textbf{Description}   \\
		\hline 
		Set of targets of opportunity (ToOs)  & \( M = \{i_1, i_2, \ldots, i_m\} \) \\
		\hline 
		Celestial coordinates of target \( i \)   & \( C_i = (\alpha_i, \delta_i) \) \\
		\hline 
		Scientific value or priority of target \( i \) & \( V_i \)  \\
		\hline
		Set of observation sites  & \( N = \{s_1, s_2, \ldots, s_n\} \)   \\
  \hline
         Geographic coordinates of site \( s \) & \( L_s = (\phi_s, \lambda_s) \)  \\
          \hline
          Set of observation bands (filters) available & \( R = \{b_1, b_2, \ldots, b_d\} \) \\
          \hline
           Required observation bands (filters) of target \( i \) & $F_i = \{f_1, f_2, \ldots, f_d\}$\\
         \hline
         Required exposure time \( e_b \) in band \( b \) for target \( i \) & \( E_i = \{(b, e_b) \mid b \in F_i\} \)  \\
         \hline
        
        Set of tasks for target $i$ based on observation mode & $K = \{j_1, j_2, \ldots, j_k\}$\\
        	\hline
         Required beginning observation time of task $j$ & $A_j$ \\
         \hline
         Scheduled beginning observation time of task $j$  & $B_j$  \\
         \hline

	\end{tabular}%
	\label{notations}%
\end{table*}%

Formally, the resource-constrained scheduling problem for follow-up observations in astronomy can be defined by a tuple $(M,N,R)$ where: $M$ represents a set of ToOs that have been identified for key follow-up observations; $N$ is a set of observation sites, which can include one or multiple geographically distributed observation sites, each equipped with multiple telescopes; $R$ is a set of resource types, which represents the types of observation bands (filters) configured at each site. Detailed parameter descriptions and notations are presented in Table \ref{notations}. We suppose to have an astronomical observation environment with $d$ types of filters. All sites have a full set of filters $R$, each telescope providing one. Tasks with different filters can overlap on the same site, maximizing telescope resource use.

\begin{table*}[htbp]
	\centering
	\caption{Properties of the incoming ToOs in follow-up observation scenarios.}
	\begin{tabular}{|c|p{2.6cm}|p{9.23cm}|}
		\hline 
		\textbf{Property ID} & \textbf{Name} & \textbf{Description}  \\
		\hline 
		\#1   & Target coordinate & Right ascension and declination are used to uniquely identify the location coordinates of the target.  \\
		\hline 
		\#2   & Filter requirement & Denote the observation filters of telescopes need to be used simultaneously to capture multi-band information.  \\
		\hline 
		\#3   & Start time & The time required to start the follow-up monitoring of the target.  \\
		\hline
		\#4   & Fade time & The time required to end the follow-up monitoring of the target.  \\
		\hline
		\#5   & Exposure time   & The requirement of the observation target for single exposure time.   \\
		\hline
		\#6   & Observation mode  & Information on whether observations need to follow immediately on from one another, or monitor a target every few minutes/hours.  \\
		\hline
		\#7   & Priority & Denote the urgency or importance of the observation target. \\
		\hline
	\end{tabular}%
	\label{properties of follow-up task}%
\end{table*}%

\begin{figure}[t]
	\centering
	\includegraphics[width=0.49\textwidth]{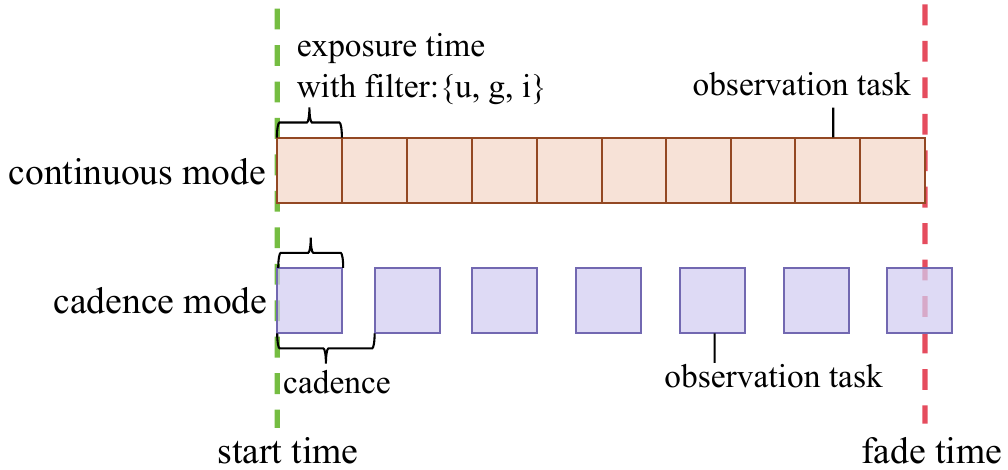} 
	\caption{Schematic diagram of the relationship between the follow-up observation target and observation task. We give the division of a target of opportunity that arises during a sky survey into multiple follow-up observation tasks (i.e., multiple exposures) according to its observational requirements in continuous observation mode or cadence observation mode. As an example, this follow-up target requires simultaneous observations in the u,g,i three bands.} 
	\label{target and task}
\end{figure}

Table \ref{properties of follow-up task} demonstrates the defined set of properties of the incoming ToOs for follow-up observations during sky survey. For target $i$, a $d$-dimensional vector $F_i$ denotes the filter requirement of the various filter types. The priority (property \#7) flags the urgency and importance of the observation target. Property \#3 and \#4 reflect the required start and end monitoring time of the follow-up observation of the target object, during which multiple shots at different times should be taken according to the required exposure time. Here, Property \#6 considers two modes of observation that are common in astronomical observation, observations of one need to follow immediately on from one another (cannot overlap), while observations of another have gaps of required cadence between them. Depending on the exposure time for target $i$ (denoted as $E_i$) required for different celestial objects, astronomers need to take multiple shots before they disappear. Each exposure of a specific sky region constitutes an observation task. Fig. \ref{target and task} shows an example of how a follow-up observation object can be divided into multiple observation tasks (i.e., exposures) based on multiple properties.

Note that the follow-up observation monitoring for target $i$, can be performed by a group of observation tasks $K$ with different numbers according to its required single exposure time. Each observation task $j$ of target $i$ can be specified as $v_j = (f_i, A_j, e_i)$. It is worth noting that the band requirements and exposure time are the same for each exposure (i.e. observation task) of each target, so $f_j$ and $E_j$ are equal to $f_i$ and $E_i$ respectively. $A_j$ denotes the required beginning observation time of task $j$. We assume that the above properties of each follow-up target is known upon arrival, and are not dependent on the site. Note that in this paper, ToOs are preprocessed, divided into observation tasks based on the observation mode. We operate under the assumption that observation tasks arrive at the telescope array in real-time, at discrete intervals. A waiting task queue is available, capable of accommodating up to $W$ tasks. When a new follow-up observation task is received, it can be promptly assigned or added to a queue. If the queue reaches its capacity, scheduling the new task requires the immediate execution of at least one task in the waiting queue to accommodate the incoming task. $W$ can be adjusted based on the practical observation scale. Additionally, we assume a fixed filter requirement throughout the entire execution of the observation tasks, with no allowance for preemption.

\subsection{Observation impact factors}
There are various of time-varying features that influence the feasibility and quality of the observation. This paper utilizes airmass \cite{kasten1989revised} as the evaluation metric to determine whether a telescope is suitable for observing a target. The airmass measures the atmospheric thickness through which astronomical light passes before reaching the telescope, which is influenced by zenith angle, altitude, atmospheric conditions, and etc. Lower airmass observations are preferred in astronomy for better image quality and less atmospheric distortion \cite{rana2019optimal}. Meanwhile, the optical radiation from the sun will also affect the resolution and clarity of the observations to some extent, so the sun's position and radiation need to be taken into account as well. The relative location of the observation site and the target, and the observation time determine the  astronomical observation conditions, which vary with time. Hence, these spatial and temporal constraints significantly increase the computational cost of resource allocation calculations.

\subsection{\textbf{MDP formulation}}

Therefore, the follow-up observation scheduling for ToOs can be modeled as an MDP with the following components.

\subsubsection{Stages}
The decision stages are the time periods at which observation scheduling decisions are made. Let $t$ denote the time period (stage) and $T$ be the set of all time periods.

\subsubsection{States}
State of the system $S_t$ encompasses all relevant information at decision stage $t$. According to the above assumptions and definitions, the states of the system at any time $t$ can be defined by: current time $t$, the current time in the scheduling horizon, current availability of observation sites, visibility window and observation quality of tasks, exposure requirements.

\subsubsection{Decisions}
At each decision stage, the following decisions need to be made (defined as the solution): target selection, which observation task $j$ to observe, which site $s$ to use for the observation, \textbf{and when to start the observation 
$B_j$}. So it can be formulated by a binary decision variable $x_{s,j,f,t} \in \{0,1\}$ indicating whether site $s$ is assigned to observe task $j$ in band $f$ at time $t$.

\subsubsection{Astronomical objectives and cost function}
Swiftly addressing the incoming follow-up observation tasks in astronomical science is crucial for timely capturing transient celestial events and phenomena. Because the rapid response enables scientists to gather critical data, facilitating real-time analysis and enhancing the chances of making groundbreaking discoveries in the dynamic and evolving cosmos. Therefore, we adopt the \textit{average task slowdown} as the primary optimization objective. Formally, for each observation task $j$ of target $i$, it can be defined as:
\begin{equation}
\eta_j=\frac{C_j-A_j}{E_i},
\end{equation} where $A_j$ and $B_j$ denote the required beginning observation time and scheduled beginning observation time, respectively. And $C_j = B_j + E_i$ is the task completion time. The objective function to minimize the sum of slowdowns for all tasks can be written as:
\begin{equation}
\min \sum_{j \in K} \eta_j = \min \sum_{j \in K} \frac{(B_j + E_i) - A_j}{E_i}
\end{equation} Normalizing the completion time by the observation task's exposure time mitigates potential bias towards lengthy observations, a situation that may arise when optimizing for objectives like mean completion time. The value of $\eta$ is $\geq$ 1. Therefore, our focus is on developing a schedule that minimizes the overall task slowdown by assigning follow-up ToO observations to telescopes equipped with the necessary filters, while adhering to the constraints of observation conditions. That means to get $B_j$ closer to $A_j$, the slowdown $\eta$ closer to 1.

The cost-to-go function $J_t(S_t)$ represents the minimum expected cost from stage $t$ to the end of the planning horizon, given the current state $S_t$. Based on the decision $x_{s,j,f,t}$, the state transition function describes how the state evolves from $S_t$ to $S_{t+1}$ can be formulated as $f(S_t,x_{s,j,f,t})$.The immediate cost $g(S_t, x_{s,j,f,t})$ represents the cost incurred at stage 
$t$ due to decision $x_{s,j,f,t}$. The cost-to-go function at stage $t$ is recursively defined as the immediate cost plus the expected cost-to-go from the next stage onward \cite{doi:10.1126/science.153.3731.34}:

\begin{equation}
J_t(S_t) = \min_{x_{s,j,f,t}} \left[ g(S_t, x_{s,j,f,t}) + \mathbb{E}[J_{t+1}(S_{t+1}) | S_t, x_{s,j,f,t}] \right]
\end{equation}Here, the immediate cost $g(S_t,x_{s,j,f,t})$ can be interpreted as the slowdown incurred by the decision at stage $t$:
\begin{equation}
g(S_t, x_{s,j,f,t}) = \eta_{i,j} = \frac{(B_{i,j} + E_{i,j}) - A_{i,j}}{E_{i,j}}
\end{equation}

So by optimizing the model, we define the scheduling problem of multi-band follow-up observation for the occurrence of ToOs in the time-domain survey of astronomical telescope arrays. The goal of the model is to minimize the task slowdown, and the cost function is described by the recursive formula of the state and decision stage, which provides a comprehensive framework for scheduling.

\section{A RL approach: \texttt{ROARS}}
\label{proposed approach}

\subsection{Graph representation}

\begin{figure*}[t]
	\centering
	\includegraphics[width=0.9\textwidth]{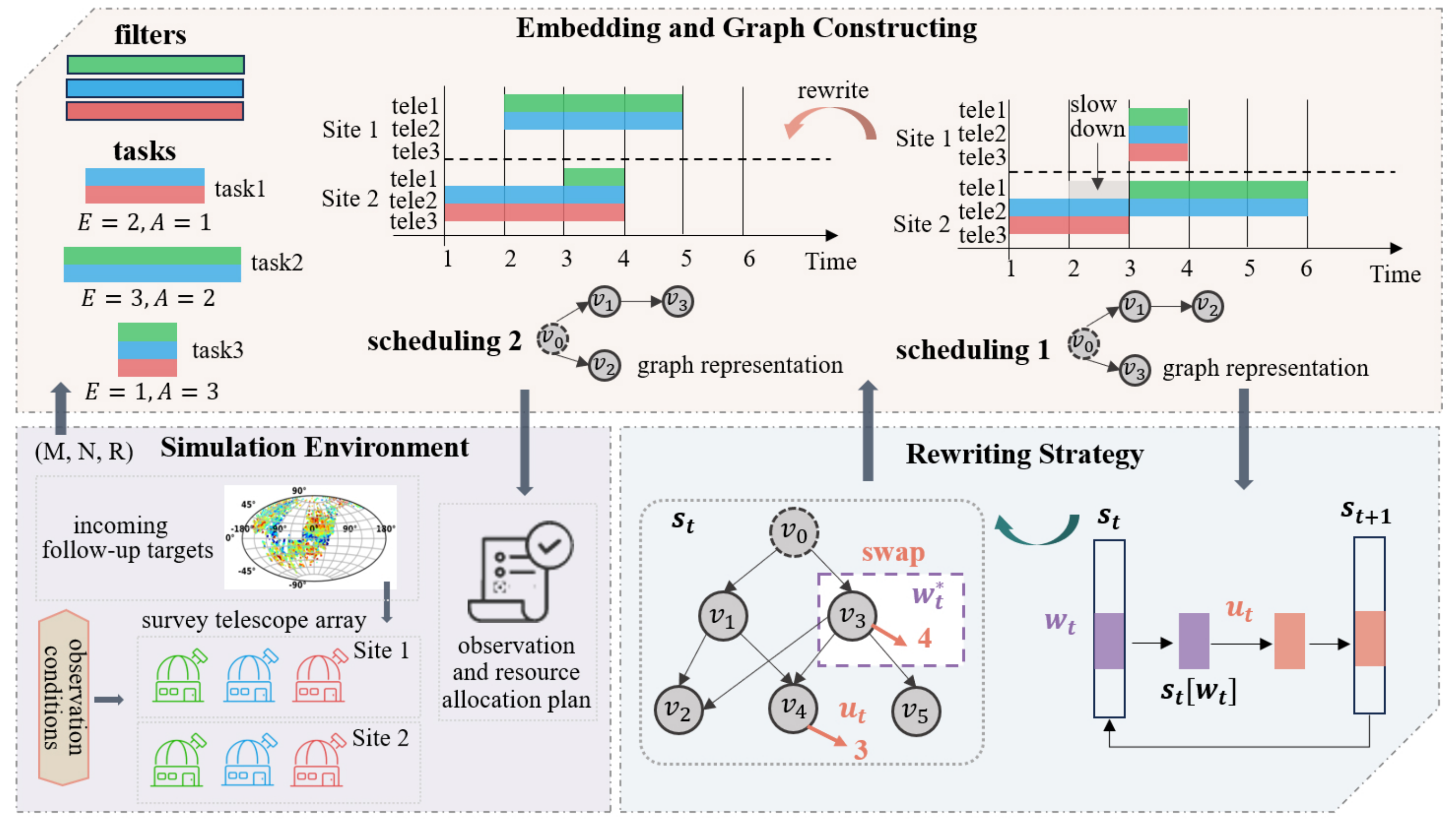} 
	\caption{An illustrative example of \texttt{ROARS}. By parsing incoming follow-up targets from the astronomical observation simulation environment, the scheduling algorithm completes resource allocation and outputs the observation plan to each observation site. We give an illustration of the rewriting optimization strategy and an example of two potential task schedules at a single observation site and their corresponding graphical representations. } 
	\label{illustration}
\end{figure*}

Based on the above problem definition, the online resource-constrained follow-up observation scheduling problem is solved by a reinforcement learning based approach named \texttt{ROARS}, depicted in Fig. \ref{illustration}. Each schedule is depicted as a DAG, illustrating the interdependence of observation task scheduling times. In this representation, each observation task $v_j$ corresponds to a node in the DAG of scheduling, with an additional node $v_0$ representing the observation telescope. If an observation task $v_j$ is scheduled upon arrival at time $A_j$ (i.e., $B_j = A_j$), we include a directed edge $\left\langle v_0, v_j\right\rangle$ in the graph. Alternatively, there must exist at least one task $v_{j^{\prime}}$ such that $C_{j^{\prime}} = B_j$ (meaning task $j$ begins immediately after task $j^{\prime}$). We include an edge $\left\langle v_{j^{\prime}}, v_j\right\rangle$ for each such task $v_{j^{\prime}}$ in the graph. 

For the follow-up observation task embedding, in intra-site telescope array setting with $D$ kinds of resources (filters), we embed each task into a vector of dimension $(D \times (E_{max} + 1) + 1)$. Here, $E_{max}$ represents the maximum exposure duration for an observation task. For distributed telescope array, it will be a $(N \times D \times (E_{max} + 1) + 1)$-dimensional vector, $N$ denotes the number of observation sites. This vector encodes details about task attributes and the observation site's status during task execution. The specifics of the task embedding are outlined below. Consider a task $v_j = (\rho_j, A_j, E_i)$ for target $i$. We represent the total resources utilization across all takes at each time step $t$ as $\rho_t^{\prime} = (\rho_{t1}^{\prime},\rho_{t2}^{\prime},\ldots,\rho_{tD}^{\prime})$. Taking intra-site observations as an example, each observation task $v_j$ is represented as a $(D \times (E_{max} + 1) + 1)$-dimensional vector, where the first $D$ dimensions of the vector are $\rho_j$, representing its observation resource requirement. The $D \times E_i$ dimensions of the vector are the concatenation of $\boldsymbol{\rho}_{B_j}^{\prime}, \boldsymbol{\rho}_{B_j+1}^{\prime}, \ldots, \boldsymbol{\rho}_{B_j+E_i-1}^{\prime}$, which describes the utilization of the observation resources during the execution of the task $v_j$. Specifically, when the energy consumption $E_i$ is less than the maximum allowed energy $E_{max}$, the subsequent $D \times (E_{max} - E_j)$ dimensions are set to zero. The final dimension of the embedding vector signifies the task's slowdown in the current schedule. Each task $v_j$ is represented by its embedding denoted as $e_j$. Additionally, the embedding of the observation telescope, denoted as $v_0$, is represented by a zero vector, $e_0 = 0$. It can be seen from Fig. \ref{illustration} that two possible observation task schedules and their corresponding graph representations may be generated. Node 0 denotes the beginning of the scheduling process, with additional instances added for multiple observation telescopes. Scheduling 2 is superior to scheduling 1 because there is no slowdown, i.e., all 3 tasks can be observed at the required observation start time (i.e. $B=A$), ensuring maximum scientific monitoring. 

We expand upon the Child-Sum Tree-LSTM architecture introduced in \cite{tai2015improved} to encode the schedule graphs.  For a job $v_j$, let $(h_1, c_1), (h_2, c_2), \ldots, (h_p, c_p)$ denote the LSTM states of all parent nodes of $v_j$, so its LSTM state can be represented as:
\begin{equation}
(h, c)=\operatorname{LSTM}\left(\left(\sum_{i=1}^p h_i, \sum_{i=1}^p c_i\right), e_j\right).
\end{equation}

\subsection{Model specification and rewriting}

\begin{algorithm}[tb]
	\caption{Algorithm of a Single Rewriting Step}
	\label{rewrite algorithm}
	\textbf{Input}: Current observation task $v_j$, another task $v_j^{\prime}$, and the dependency graph representation of task schedule $s_t$ \\
	\textbf{Output}: The dependency graph representation of task schedule in next time step $s_{t+1}$
	\begin{algorithmic}[1] 
		
		\IF{$C_{j^\prime} < A_j \OR C_{j^\prime} == B_j $} 
		\RETURN $s_t$
		\ENDIF
		
		\IF {$j^{\prime} \neq 0 $} 
		\STATE $B_j^{\prime} = C_{j^{\prime}}$ 
		\ELSE 
		\STATE $B_j^{\prime} = A_j$
		\ENDIF
		
		\STATE $C_j^{\prime} = B_j^{\prime} + E_j$
		\STATE $J = $ all tasks in $s_t$ except $v_j$ that are scheduled within $\left[B_j^{\prime}, C_j^{\prime}\right]$
		
		\STATE Sort $J$ in the topological order
		
		\FOR{ $v_p \in J$}
		\STATE $B_p^{\prime} = $ the earliest time that task $v_p$ can be scheduled 
		\STATE $C_p^{\prime}=B_p^{\prime}+E_p$
		\ENDFOR
		
		\FOR{$v_p \notin J, B_p^{\prime}=B_p, C_p^{\prime}=C_p$}
		
		\STATE $s_{t+1}=\left\{\left(B_p^{\prime}, C_p^{\prime}\right)\right\}$
		\ENDFOR
		\RETURN $s_{t+1}$
	\end{algorithmic}
\end{algorithm}

After constructing the graph representation, we train a neural-based policy to iteratively refine the current scheduling solution by locally rewriting parts of it until convergence. This approach draws inspiration from \cite{chen2019learning}. We employ the end-to-end reinforcement learning to train the policy, encouraging the cumulative enhancement of the solution. For the telescope observation scheduling problem, finding a feasible solution that meets the constraints of observation time and geographical location is straightforward. Additionally, the search space displays favorable local structures that facilitate incremental enhancements to the solution. Thus, a comprehensive solution offers a contextual basis for enhancement through a rewriting-based approach, facilitating the computation of additional features, a challenge when generating a solution from scratch. Various solutions may converge towards optimization through a shared pathway, which could be encapsulated as local rewriting rules. Furthermore, straightforward rules such as task swapping could enhance performance. These aspects enable the application of the rewriting formulation to diverse instances of follow-up observation scheduling. We can train the neural network to investigate relationships among diverse solutions within the search space. Our rewriting strategy incorporates a region-selection policy and a rule-selection policy.

Each solution represents a state, and every local region, along with its corresponding rewriting rule, acts as an action. Algorithm \ref{rewrite algorithm} describes steps for a single rewriting. The rewriting rules involve relocating the present task $v_j$ to be positioned as a child of another job $v_{j^{\prime}}$ or $v_0$ within the graph. This results in scheduling observation task $v_j$ to commence either after task $v_{j^{\prime}}$ concludes or at its arrival time $A_j$. As depicted in Fig. \ref{illustration}, $s_t$ represents the dependency graph of the observation task schedule. Each circle with an index greater than 0 denotes a task node, while node 0 serves as an additional representation of the observation site. The graph's edges denote the observation dependencies among follow-up tasks. The region-picking policy chooses a task $\omega_t$ from all task nodes for rescheduling, while the rule-picking policy determines a movement action $u_t$ for $\omega_t$. Afterwards, $s_t$ is modified to obtain a new dependency graph $s_{t+1}$.

Let $\mathcal{U}$ be the rewriting rule set, shown in Fig. \ref{illustration}. Assume that $s_t$ represents the current solution (or state) at iteration $t$. Firstly, a state-dependent region set $\Omega\left(s_t\right)$ is computed, which is problem-dependent, and covers all follow-up observation task nodes for scheduling. We then select a region $\omega_t \in \Omega\left(s_t\right)$ using the region-picking policy $\pi_\omega\left(\omega_t \mid s_t\right)$. For each $\omega_t \in \Omega\left(s_t\right)$, we calculate a score $Q(s_t, \omega_t)$, which reflects the potential benefit of rewriting. A higher score suggests that rewriting $s_t\left[\omega_t\right]$ may be advantageous.

Afterwards, a rewriting rule $u_t$ is selected for the region $\omega_t$ using the rule-picking policy $\pi_u\left(u_t \mid s_t\left[\omega_t\right]\right)$, where $s_t\left[\omega_t\right]$ denotes a subset of the state $s_t$. The chosen rewriting rule $u_t \in \mathcal{U}$ is then applied to $s_t\left[\omega_t\right]$, resulting in the subsequent state represented as $s_{t+1} = f\left(s_t, \omega_t, u_t\right)$. The rewriting sequence in the forward pass can be denoted as 
\begin{equation}
    s_T = \left(s_0,\left(\omega_0, u_0\right)\right),\left(s_1,\left(\omega_1, u_1\right)\right), \ldots,\left(s_{T-1},\left(\omega_{T-1}, u_{T-1}\right)\right). 
\end{equation}

Hence, commencing with an initial solution (or state) $s_0$, our aim is to discover a sequence of rewriting steps $s_T$ that minimizes the final cost $c(s_T)$.

Note that we both use fully connected neural networks for the prediction of region score and selection of a rewriting rule.

\subsection{Training details}
Our region-picking policy $\pi_\omega$ and rule-picking policy $\pi_u$ are trained in the meantime. The reward function of training can be defined as $r\left(s_t,\left(\omega_t, u_t\right)\right)=c\left(s_t\right)-c\left(s_{t+1}\right)$. For $\pi_\omega$, we express the parameterization as a softmax function of the $Q\left(s_t, \omega_t ; \theta\right)$:
\begin{equation}
\pi_\omega\left(\omega_t \mid s_t ; \theta\right)=\frac{\exp \left(Q\left(s_t, \omega_t ; \theta\right)\right)}{\sum_{\omega_t} \exp \left(Q\left(s_t, \omega_t ; \theta\right)\right)}
\end{equation}The training of $Q\left(s_t, \omega_t ; \theta\right)$ involves aligning it with the cumulative reward obtained from the present learning policies $\pi_\omega$ and $\pi_u$:

\begin{equation}
L_\omega(\theta)=\frac{1}{T} \sum_{t=0}^{T-1}\left(\sum_{t^{\prime}=t}^{T-1} \gamma^{t^{\prime}-t} r\left(s_t^{\prime},\left(\omega_t^{\prime}, u_t^{\prime}\right)\right)-Q\left(s_t, \omega_t ; \theta\right)\right)^2
\end{equation} 
Here, $T$ represents the episode length, indicating the count of rewriting steps, while $\gamma$ signifies the decay factor. Regarding the rule-picking policy, we employ the advantage actor-critic mechanism, leveraging $Q\left(s_t, \omega_t ; \theta\right)$ as the critic. This approach mitigates bootstrapping-related issues that may arise from sample insufficiency and training instability. The advantage function can be represented as: \begin{equation}
\Delta\left(s_t,\left(\omega_t, u_t\right)\right) \equiv \sum_{t^{\prime}=t}^{T-1} \gamma^{t^{\prime}-t} r\left(s_t^{\prime},\left(\omega_t^{\prime}, u_t^{\prime}\right)\right)-Q\left(s_t, \omega_t ; \theta\right)
\end{equation} 
The loss function of the rule selector can be represented as:
\begin{equation}
L_u(\phi)=-\sum_{t=0}^{T-1} \Delta\left(s_t,\left(\omega_t, u_t\right)\right) \log \pi_u\left(u_t \mid s_t\left[\omega_t\right] ; \phi\right)
\end{equation}
Moreover, we denote the overall loss function as  $L(\theta, \phi)=L_u(\phi)+\alpha L_\omega(\theta)$, where $\alpha$ serves as a hyperparameter.

During the training process, $\alpha$ is set to 10. The numbers of region picking and rule picking are both 15 for intra-site telescope array, while both 30 for distributed array, which are sufficient for figuring out a competitive scheduling solution. The hyperparameter specifying the total number of rewriting steps is set to 100 iterations. For all tasks in our evaluation, suppose the probability of re-sampling the region for rewriting is $1-p_c$, $p_c$ starts with a initial value of 0.5, and is gradually decayed by 0.8 every 1000 time steps until it reaches a minimum value of 0.01, at which point it remains constant. We set the decay factor for the cumulative reward to $\gamma=0.9$, and the initial learning rate to 1$e$-4, which is then decayed by a factor of 0.9 every 1000 time steps. Furthermore, we maintain a fixed batch size of 128 during training. The model is optimized using the Adam, with all weights initialized uniformly randomly within the range of $[-0.1, 0.1]$.

\section{Experiment}
The tested algorithms were implemented in Python. In our evaluation, the neural networks are implemented using PyTorch \cite{paszke2017automatic}. All implementation and experiments were performed on an Ubuntu server featuring a 4-core Intel Xeon CPU (clocked at 2.2 GHz), 32 GB of memory, and a Tesla V100 GPU.

\subsection{Setup}

\begin{figure*}[t]
	\centering
	\includegraphics[width=0.9\textwidth]{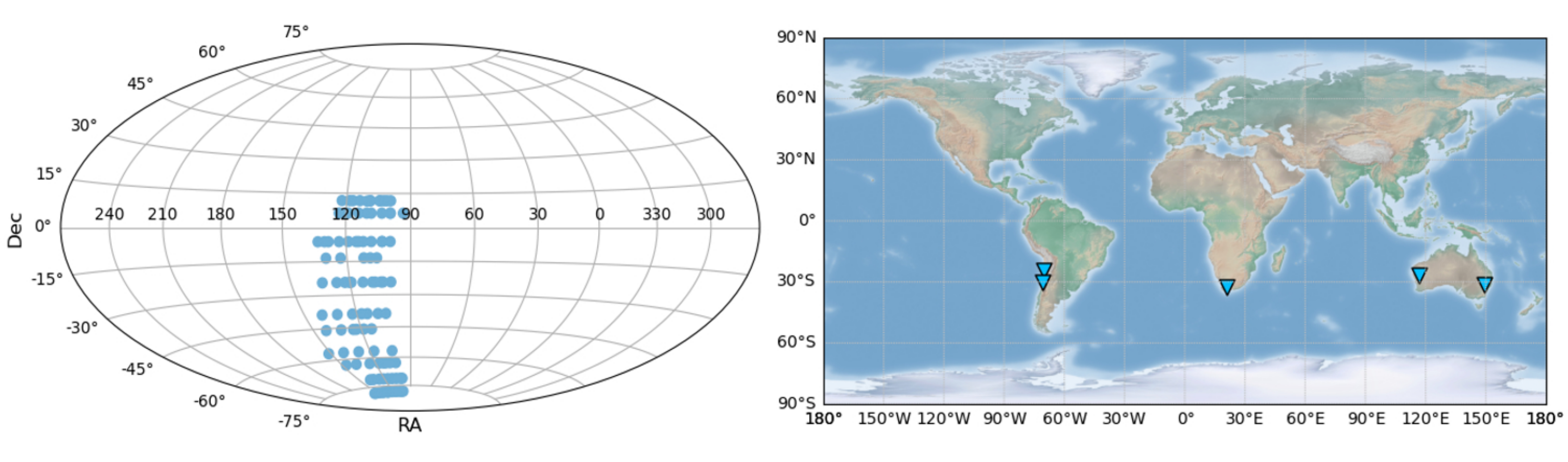} 
	\caption{The location information of observation sites and fields in dataset generation.  } 
	\label{sites and fields}
\end{figure*}

We conduct experiments on simulated data based on various scenarios in real-life settings to investigate the effectiveness of \texttt{ROARS} in terms of solution quality, computational speed, robustness and scalability. We develop a simulator environment to model the observations of both intra-site and distributed telescope arrays, including the diverse conditions, telescope states, and follow-up targets for telescope arrays to test the algorithm under varied settings. These conditions act as constraints, aiding in determining available resources for a target at a given time. \textbf{As shown in Fig. \ref{sites and fields}, we select 5 real observation sites worldwide to simulate the formation of the telescope array. The position of the ToOs are generated from 100 sky fields with the configuration parameters described later. In addition, we simulate the arrival of ToOs within 4 hours. Since different ToOs have varying observation requirements and need to be monitored over a period of time, the generated instances are based on a collection of observation tasks according to the duration and exposure time required by the ToOs. }The simulator serves for both model training and evaluation. Note that the coordinate of the targets and observation sites are collected from real observations. According to the configuration of the real telescope array under construction, we use $u$, $g$ and $i$ bands as possible filter requirement inputs. In addition, We generated one hundred thousand follow-up observation task sequences randomly, allocating 80\% for training and reserving 10\% each for validation and testing. The waiting task queue length, denoted as $W$, is fixed at 10. Initial schedules were generated using the \textit{First Come First Serve} (\texttt{FCFS}) method, known for its low overhead during construction.

\subsubsection{Evaluation metric.} According to the actual astronomical observation needs, we utilize the \textit{average task slowdown} $\eta_j \equiv\left(C_j-A_j\right) / E_j$ as the evaluation metric. It is preferred that follow-up observation tasks be handled as soon as possible after arrival.

\subsubsection{Task properties.}
In order to adequately test the robustness and generalization of 
\texttt{ROARS}, various observation task properties are evaluated: (1) \textit{Average arrival rate of ToOs}: the probability of a new ToO arrival, the \textit{Steady} task frequency sets it to be 10\% (because normally targets that need to be followed up in the sky survey observation are in the minority), and \textit{Dynamic} task frequency  indicates that the ToO arrival rate varies randomly at each time step; (2) \textit{Duration of the follow-up observation}: the time from the beginning when the opportunity target requires observation monitoring to the end, \textit{Long} for the time duration requirement of the target is in $[120, 240]$ minutes, \textit{Short} for $[60,119]$ minutes, and \textit{Non-uniform} task duration; (3) \textit{Resource distribution}: observation tasks might have different resource requirements, we consider \textit{Uniform} resource as the ToO selects two of the three possible resources with the same probability to observe simultaneously, while \textit{Non-uniform} refers to simultaneous observations of one, two or three of bands required with 10\%, 20\% and 30\% probability, respectively; (4) \textit{Single exposure time}: length of each observation task, \textit{Long} means single exposure time is in $[10,20]$ minutes, \textit{Short} for $[1,9]$ minutes, and \textit{Non-uniform} exposures; (5) \textit{Observation mode}: the two ways to make a series of sequential observations in astronomy, one is to perform observations that follow immediately on from one another (\textit{Exposure count}), the other is to monitor the target with required \textit{Cadence}.

\subsubsection{Baselines on heuristics.}
For intra-site telescope array scenarios, we implement 5 online heuristic approaches that are popular in existing machine scheduling and RCPSP problems for comparison. \textit{Shortest Task First} (\texttt{STF}) allocates the tasks with the shortest exposure time in the waiting task queue at each time step. \textit{First Come First Serve} (\texttt{FCFS}) schedules each observation task in the increasing order based on the arrival time. \textit{Earliest Due Date} (\texttt{EDD}) schedules the observation tasks with the earliest end time. \textit{Shortest Processing Time} (\texttt{SPT}) prioritizes targets with the shortest duration (from the time required to start monitoring to its fade) to be monitored. And \textit{Resource Intensity Priority} schedules the target with the highest resource requirements first. In this paper, resource demand intensity is defined as the number of required observation resource types multiplied by the total exposures. In addition, we have also tried to employ the optimal solver (e.g. Gurobi \cite{optimization2014inc} and CBC \cite{forrest2005cbc}) for the problem in this context, but both seem intractable in general. Therefore, we omit the comparison.

For scenarios of distributed observation sites, heuristically obtaining a feasible solution is more complicated. It involves two steps of selecting the follow-up task and selecting the site, so we correspondingly design the following baselines. We make heuristic site selections based on \textit{equipment priority factor} and \textit{observation quality}, which are what astronomers tend to do in the current study. In practical observations, the \textit{equipment priority factor} for each site is usually related to the weather changes, which can be predicted or obtained in real time. Here we use the airmass of each schedule to represent the \textit{observation quality}, while generate the \textit{equipment priority factor} of each site randomly as input. It should be noted that \texttt{ROARS} has not been specifically optimized for these two parameters. The heuristic baselines are \textit{Shortest Task best Quality site First} (\texttt{SQTF}), \textit{Shortest Task best Priority site First} (\texttt{SPTF}), \textit{First come Task best Quality site First} (\texttt{FQTF}), \textit{First come Task best Priority site First} (\texttt{FPTF}), \textit{shortest Processing best Quality site First} (\texttt{PQTF}), \textit{shortest Processing best Priority site First}(\texttt{PPTF}), \textit{earliest Due Task best Quality site First} (\texttt{DQTF}), \textit{earliest Due Task best Quality site First} (\texttt{DPTF}), \textit{Shortest Task best Quality site First} (\texttt{RQTF}), and \textit{Shortest Task best Quality site First} (\texttt{RPTF}).

\subsubsection{Baselines on offline scheduling.} In order to evaluate the effectiveness of these algorithms, we examine an offline scenario where the complete task sequence is known beforehand. This is equivalent to simulating an unbounded waiting task queue. This setting, with additional prior knowledge, serves as a strong baseline for evaluation. We utilize \texttt{STF-offline}, a basic heuristic approach that schedules tasks in ascending order of duration.

\subsection{Result comparison}







\begin{table}[htbp]
	\centering
	\caption{Statistics of solution quality compared to baselines in terms of average slowdown and computation time. }
	\begin{tabular}{lcc}
		\hline
		& Avg. slowdown & Time (s) \\
		\hline
		STF   & 16.47 & 0.11 \\
		
		FCFS   & 27.23 & 0.01 \\

        EDD & 23.59 &   0.12 \\

        SPT & 15.49 &  0.08 \\
 
        RIP &25.93  &  0.13 \\
		
		Offline & 7.82  & 0.19 \\
		
		\texttt{ROARS} &  7.34     &0.08  \\
		\hline
	\end{tabular}%
	\label{single site comparison table}%
\end{table}%

Our first experiment compares the solution quality in terms of the average slowdown and computation time, as demonstrating in Table \ref{single site comparison table}. In this experiment, training and testing instances are with steady task frequency, observation mode with cadence, non-uniform duration of follow-up task observation (20\% long, 80\% short), non-uniform resource distribution, and non-uniform single exposure time (20\% short, 80\% long). We conduct the experiments on one thousand different instances and average the results. It can be seen that \texttt{ROARS} outperforms both online heuristic baselines and the offline approach. It is more time-efficient than \texttt{STF}, \texttt{EDD}, and \texttt{Offline}. For the solution quality, \texttt{ROARS} can achieve nearly a 50\% improvement compared to \texttt{STF} and \texttt{SPT}, which are the better performing heuristics. The average slowdown of \texttt{Offline} approach is comparable to that of \texttt{ROARS}, but is much more time-consuming. The \texttt{Offline} approach takes the knowledge of the entire incoming task sequence into account, which is helpful but costs larger time for analysis.

\subsubsection{Results on generalization of various distributions.}

\begin{figure*}
	\centering
	\subfigure[]{\includegraphics[width=2.62in]{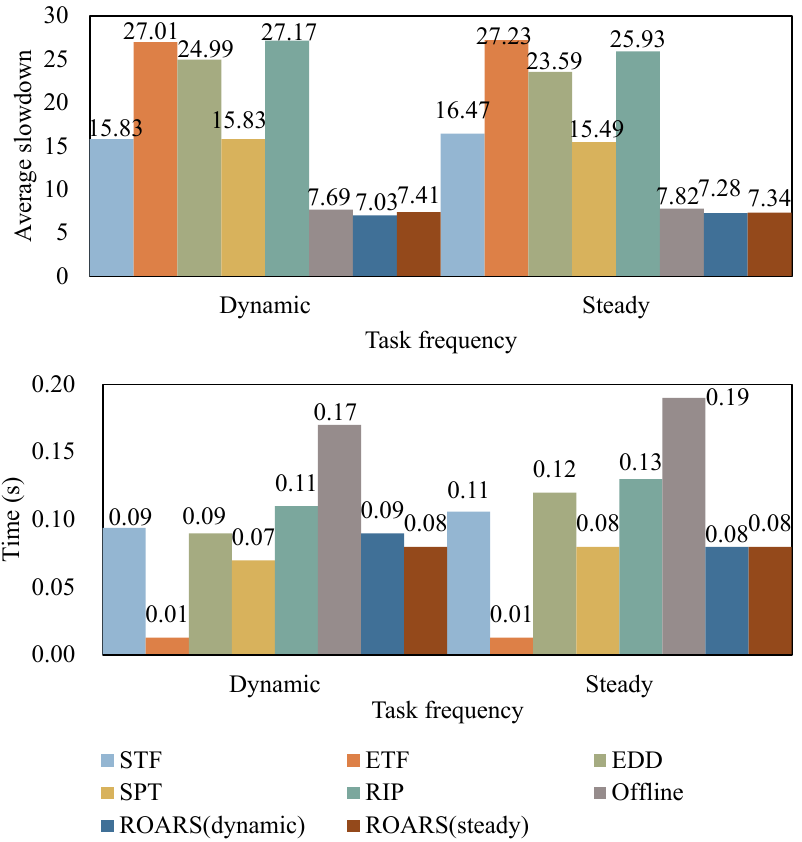}}
	\subfigure[]{\includegraphics[width=3.62in]{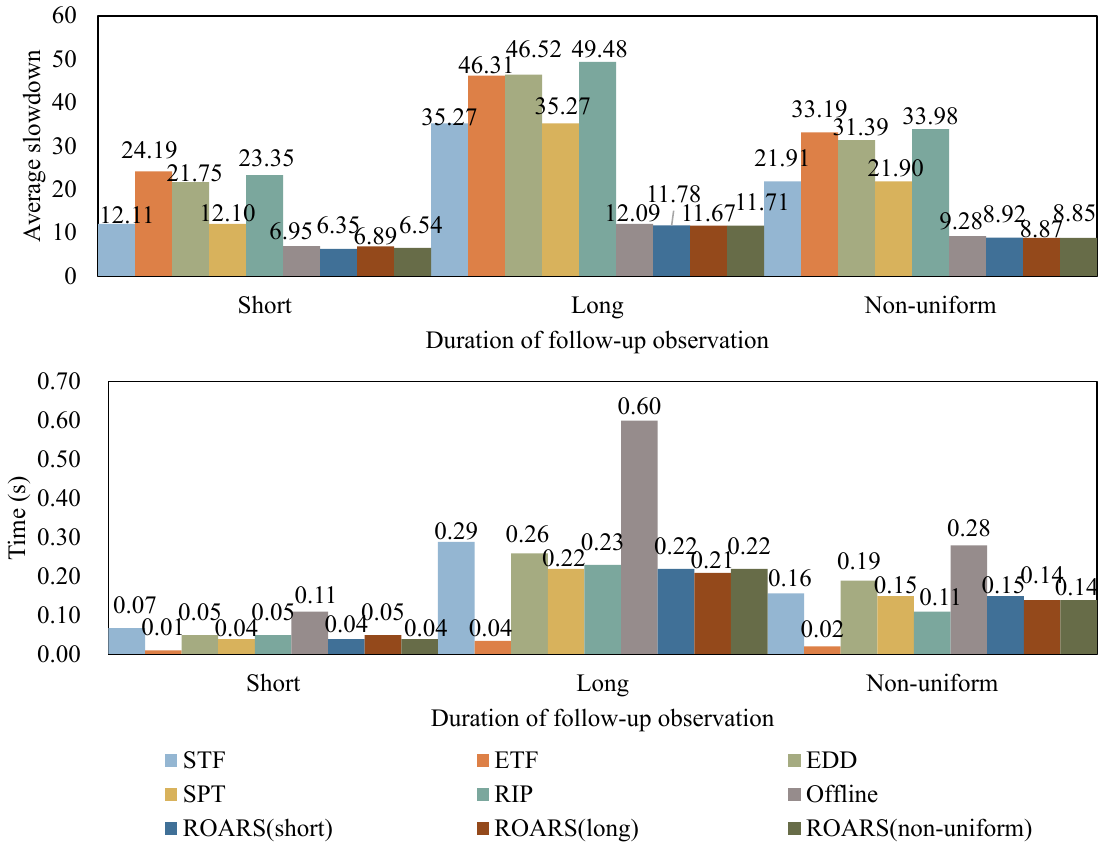}}
	\subfigure[]{\includegraphics[width=2.61in]{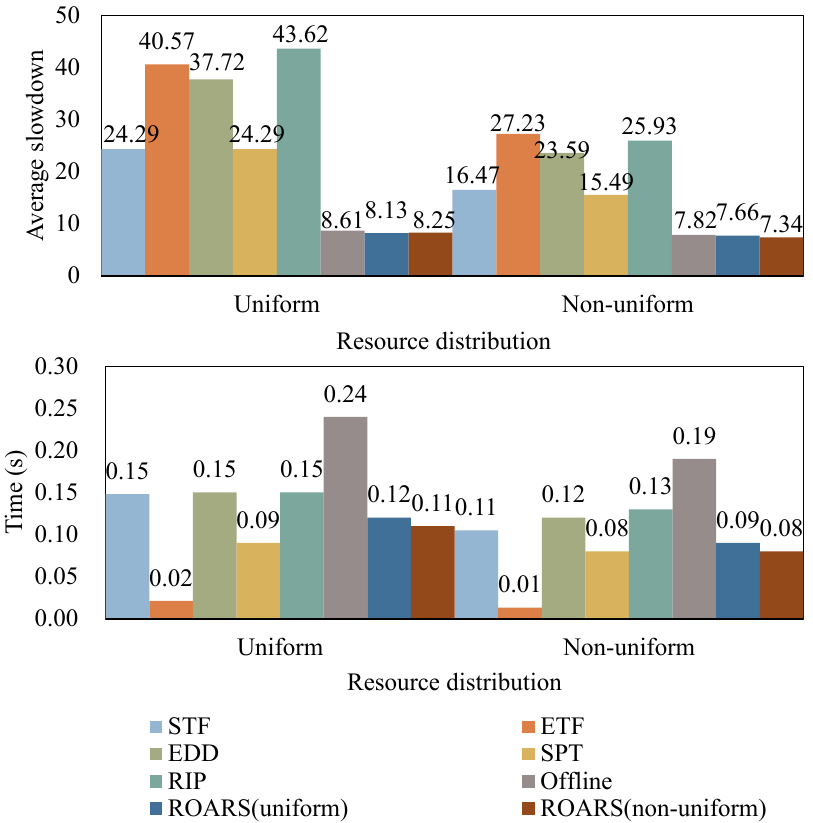}}
	\subfigure[]{\includegraphics[width=3.59in]{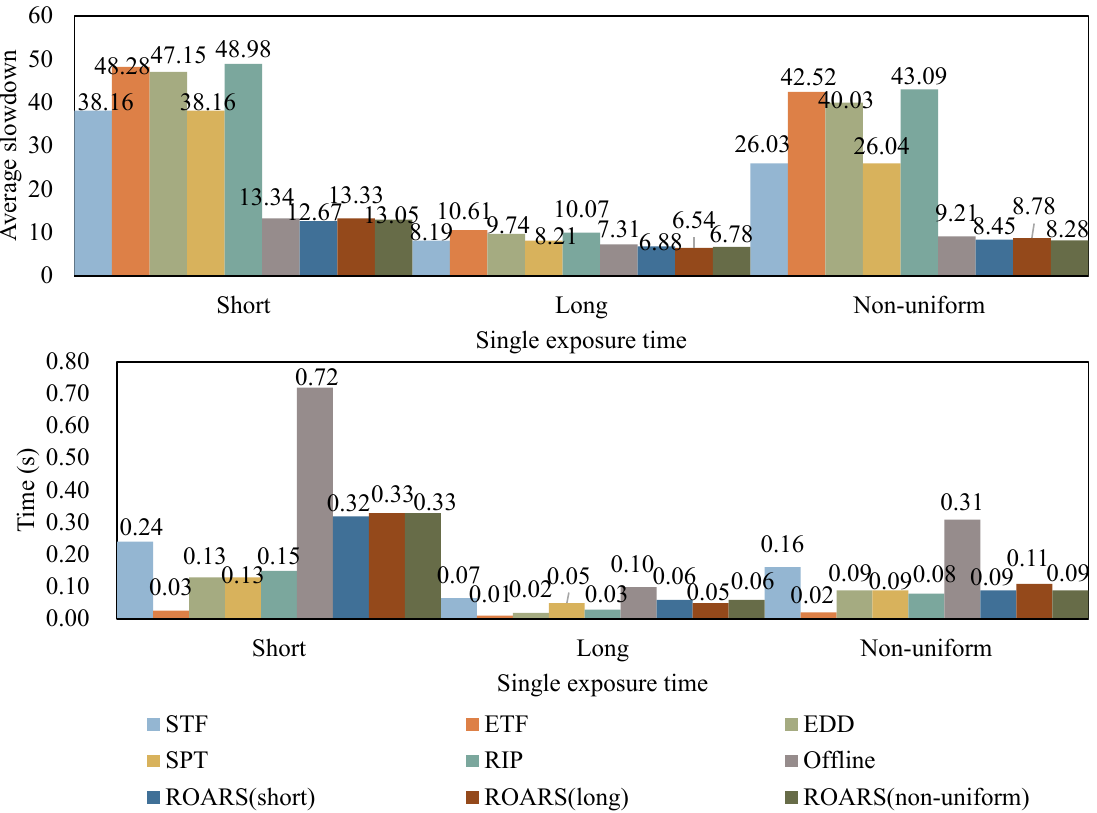}}
	\caption{Experimental results of \texttt{ROARS} varying the following task properties: (a) task frequency of the incoming ToOs; (b) duration of follow-up observation; (c) resource distribution; and (d) single exposure time. Except for the influence of the properties to be explored on the results, the remaining properties in each experiment are set to be the same as those in the experiment presented in Table \ref{single site comparison table}.}
	\label{quality comparisons}
\end{figure*}

Our second experiment evaluates the proposed \texttt{ROARS} on different distributions of the incoming ToO observation tasks. As shown in Fig. \ref{quality comparisons}, we perform ablation experiments in different scenarios for each property. We can observe that for the average slowdown, \texttt{ROARS} is superior to all various input distributions. When we set all single exposure time of observation tasks long, \texttt{STF} and \texttt{Offline} show acceptable performance, similar to \texttt{ROARS}. This is  because for the same observation task duration, the longer the single exposure time, the fewer observations need to be performed, and the difficulty of solving the calculation decreases. Therefore, it illustrates that the proposed \texttt{ROARS} can deal with more complicated scenarios and the solutions obtained are more effective when extended to larger scale problems. In addition, \texttt{ROARS} can exhibit robust generalization to distributions not encountered during training, demonstrating the efficacy of local rewriting rules. Leveraging local context proves advantageous, yielding solutions with broader applicability, aligning with our design principles.

\begin{figure*}[t]
	
	\includegraphics[width=0.78\textwidth]{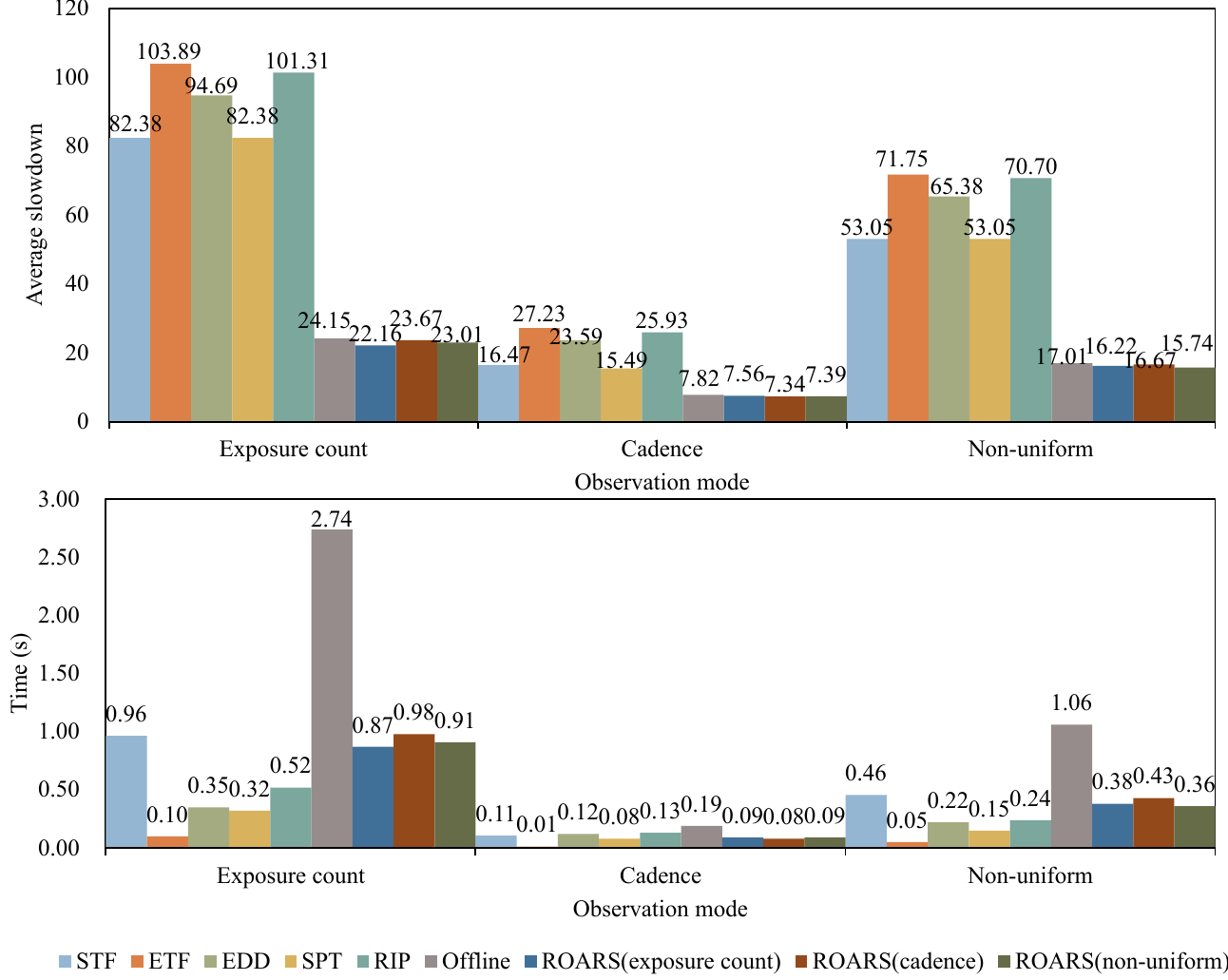} 
	\centering
	\caption{Experimental results of \texttt{ROARS} varying the observation modes.}
	\label{observation mode}
\end{figure*}

Furthermore, Fig. \ref{observation mode} demonstrates the effectiveness and robustness of \texttt{ROARS} when generalizing to different observation modes. In the field of astronomical observation, due to the rotation of the Earth, the effect of tasks implementing observations at different times varies greatly. The experimental results indicate that our proposed \texttt{ROARS} can effectively adapt to different observation modes and efficiently generate scheduling schemes based on observable times.

\subsubsection{Extended to distributed telescope array.}

\begin{table}[htbp]
	\centering
	\caption{Statistics of solution quality compared to baselines in distributed observation environment. Results are compared in terms of average slowdown and computation time.}
	\begin{tabular}{lccc}
		\hline
		& Avg. slowdown & Time (s)  \\
		\hline
		SQTF  & 11.75 & 0.24   \\
		
		FQTF  & 23.28 & 0.13  \\
		
		SPTF  & 13.75 & 0.22  \\
		
		FPTF  & 20.91 & 0.03 \\

        PQTF  &14.75 &0.09\\

        PPTF  &16.67 &0.09 \\

        DQTF   &23.77 &0.12\\

        DPTF   &24.89 &0.14\\

        RQTF  &24.00  &0.15\\

        RPTF    &26.02 &0.18\\

		Offline & 5.16 & 1.26  \\
		
		\texttt{ROARS} &   4.89    &   0.22    \\
		\hline
	\end{tabular}%
	\label{distributed site comparison}%
\end{table}%

Moreover, we extend the \texttt{ROARS} to the environment of distributed telescope array observation. This means that for the same observation target, the observation sites have appropriate observation resources at different times. The same setting for the observation task properties are adopted as experiments for single site scenario. Table \ref{distributed site comparison} presents the results comparisons to heuristic algorithms (including selecting tasks and sites heuristically). Due to more sufficient observation resources, the overall slowdown is reduced compared with results of the intra-site telescope array observations, indicating that multiple sites cooperate to process the coming follow-up tasks. On the contrary, distributed observation resources also increase the computational complexity of the problem, resulting in longer computing time. It can be observed that the \textit{Shortest Task First} heuristics lead to a better overall slowdown than the other heuristics, while \textit{Resource Intensity Priority} performs the worst. This may be due to the different time windows and visibility constraints of different observation targets, leading to resource competition and saturation. Prioritizing tasks with high resource intensity may not be able to complete other tasks in the optimal time window, resulting in a decrease in overall efficiency. Compared with the static heuristic method, our DRL-based method can constantly adjust the strategy according to the real-time observation data during the learning process to adapt to the changes in the environment and the dynamic needs of tasks, and optimize the long-term returns. Our results improved by 5.2\% compared to offline algorithms that presume knowledge of the entire task sequence, further underscoring the effectiveness of \texttt{ROARS}. Note that the modeling of site priority and its impact on distributed scheduling results will be further explored.

\section{Conclusion}

In this paper, we have presented \texttt{ROARS}, an online scheduling approach for resource-constrained follow-up observation problem in astronomy. Our approach relies on modeling each schedule as a DAG which is encoded using extended Child-Sum Tree-LSTM architecture, and iteratively refining an existing solution towards optimality using deep reinforcement learning. Our proposed algorithm is validated and proven effective through numerical simulations conducted with real-world scenarios. Experimental results show that \texttt{ROARS} can infer schedules on unseen instances of higher quality than those produced by popular heuristics and even the offline setting, in various astronomical observation settings.

Furthermore, we will investigate the enhancement of solving capabilities by improving the model structure, especially in how to learn implicit competition for observation targets between the distributed observation sites. Moreover, now we use fully connected neural networks for region selection and rule selection, providing the flexibility to replace them with more advanced neural network models for further performance enhancement. \texttt{ROARS} can also be extended to other complex variants of dynamic resource management problems in astronomical observation domain, such as multi-objectives, more complex observation modes, etc. Our approach is designed for the ongoing deployment of the global telescope array for sky survey observations, serving as a crucial component in this initiative. Further experimentation is currently underway for the integration and refinement into the practical observation environment.

\section*{CRediT authorship contribution statemen}
\textbf{Yajie Zhang}: Conceptualization, Writing – original draft, Writing – review \& editing, Visualization, Software, Validation. \textbf{Ce Yu}: Supervision, Project administration, Funding acquisition. \textbf{Chao Sun}: Methodology, Validation, Funding acquisition. \textbf{Jizeng Wei}: Funding acquisition, Writing – review \& editing. \textbf{Junhan Ju}: Writing – review \& editing, Visualization, Data curation, Validation. \textbf{Shanjiang Tang}: Formal analysis, Methodology, Writing – review \& editing.

\section*{Declaration of competing interest}
The authors declare that they have no known competing financial interests or personal relationships that could have appeared to
influence the work reported in this paper.

\section*{Data availability}
Domain models and problem instances used in this paper can be provided by the authors upon request.

\section*{Acknowledgements}
This work was financially supported by National Key R\&D Program of China No. 2023YFA1608301 and the National Natural Science Foundation of China (NSFC) No. 12133010 and No. 12273025.







\bibliographystyle{elsarticle-num} 
\bibliography{./reference.bib}

\end{document}